\pdfoutput=1

\documentclass[11pt]{article}

\usepackage[]{acl}

\usepackage[T1]{fontenc}

\usepackage[utf8]{inputenc}

\usepackage{microtype}
\usepackage{times}
\usepackage{latexsym}
\usepackage{amssymb}
\usepackage{amsfonts}
\usepackage{amsmath} 
\usepackage{amsthm} 
\usepackage{booktabs}
\usepackage{enumerate}
\usepackage{graphicx}
\usepackage{subfigure}
\usepackage{xspace}
\usepackage{float}
\usepackage{bbm}
\usepackage{bm}
\usepackage{multirow}
\usepackage{booktabs}
\usepackage{color}
\usepackage{framed}
\usepackage{stfloats}
\usepackage{iitem}
\usepackage{makecell}
\usepackage{float}
\restylefloat{table}
\usepackage{placeins}
\usepackage{afterpage}
\usepackage[normalem]{ulem}
\useunder{\uline}{\ul}{}
\usepackage{url}
\newcommand{\paratitle}[1]{\vspace{1.5ex}\noindent\textbf{#1}}
\newcommand{\ie}{\emph{i.e.,}\xspace}

\newcommand{\eg}{\emph{e.g.,}\xspace}

\newcommand{\ignore}[1]{}
\newcommand{\tabincell}[2]{\begin{tabular}{@{}#1@{}}#2\end{tabular}}

\usepackage{xcolor}
\definecolor{tOrange}{RGB}{255,165,0}
\definecolor{tBlue}{RGB}{24,116,205}
\definecolor{tPink}{RGB}{255,20,147}
\definecolor{tGreen}{RGB}{50,205,50}
\definecolor{tGold}{RGB}{255,215,0}

\title{\textsc{Elmer}: A Non-Autoregressive Pre-trained Language Model for Efficient and Effective Text Generation}

\author{
	Junyi Li\textsuperscript{\rm{1,3,4}}, 
	Tianyi Tang\textsuperscript{\rm{1}},
        Wayne Xin Zhao\textsuperscript{\rm{1,4}\thanks{\ \ Corresponding author}\ }, 
	Jian-Yun Nie\textsuperscript{\rm{3}} {\rm and} 
	Ji-Rong Wen\textsuperscript{\rm{1,2,4}} \\
	\textsuperscript{1}Gaoling School of Artificial Intelligence, Renmin University of China \\
	\textsuperscript{2}School of Information, Renmin University of China \\
	\textsuperscript{3}DIRO, Universit\'{e} de Montr\'{e}al \\
	\textsuperscript{4}Beijing Key Laboratory of Big Data Management and Analysis Methods \\
	\texttt{\{lijunyi,steven\_tang\}@ruc.edu.cn} \quad
	\texttt{batmanfly@gmail.com} \\
}

\begin{document}
\maketitle

\begin{abstract}
We study the text generation task under the approach of pre-trained language models (PLMs). Typically, an auto-regressive (AR) method is adopted for generating texts in a token-by-token manner.  
Despite many advantages of AR generation, it usually suffers from inefficient inference. Therefore, non-autoregressive (NAR) models are proposed to generate all target tokens simultaneously. However, NAR models usually generate texts of lower quality due to the absence of token dependency in the output text. In this paper, we propose \textbf{\textsc{Elmer}}: an \textbf{\underline{E}}fficient and effective P\textbf{\underline{LM}} for NAR t\textbf{\underline{E}}xt gene\textbf{\underline{R}}ation to explicitly model the token dependency during NAR generation. By leveraging the early exit technique, \textsc{Elmer} enables the token generations at different layers, according to their prediction confidence (a more confident token will exit at a lower layer). 
Besides, we propose a novel pre-training objective, Layer Permutation Language Modeling, to pre-train \textsc{Elmer} by permuting the exit layer for each token in sequences. Experiments on three text generation tasks show that \textsc{Elmer} significantly outperforms NAR models and further narrows the performance gap with AR PLMs (\eg \textsc{Elmer} (29.92) vs BART (30.61) ROUGE-L in XSUM) while achieving over 10 times inference speedup.

\end{abstract}

\section{Introduction}
\label{sec-intro}


Since the advant of GPT-2~\cite{gpt2}, pre-trained language models (PLMs) have achieved 
state-of-the-art performance across text generation tasks, which aim to generate human-like texts on demand~\cite{gpt3,PLM4TG_survey}. These PLMs usually adopt an \textit{auto-regressive} (AR) fashion to generate texts token-by-token: the next token is predicted based on all previously generated tokens. A major limitation of this approach is that it is hard to be parallelized for the inference process, thus leading to a relatively high inference latency~\cite{Gu0XLS18}.
Such a limitation prevents AR models from wide deployment in online real-time applications, such as query rewriting in search engines and online chat-bot. Moreover, AR models are prone to suffering from the exposure bias problem since there is a gap between AR training and inference~\cite{DBLP:conf/acl/ZengN21}. These concerns have sparked extensive interests in \textit{non-autoregressive} (NAR) models for text generation~\cite{Gu0XLS18}. 


Compared to AR models, NAR models predict target tokens in all positions simultaneously and independently~\cite{Gu0XLS18}. This full parallelism leads to an efficient and low-latency inference process. However, the independence assumption prevents NAR models from learning the dependency among target tokens, resulting in accuracy degradation~\cite{dad}. One widely-used solution to improve the NAR generation quality is to iteratively refine outputs~\cite{GuWZ19,GhazvininejadLL19}, which however leads to the loss in the speed-up advantage. In addition, many studies aim to learn the input-output mapping for more accurate generation via embedding mapping~\cite{GuoTHQXL19}, latent alignment~\cite{LibovickyH18}, and discrete variables~\cite{MaZLNH19}. While easing the difficulty of NAR generation to some extent, these methods still struggle for generating complex sentences. Therefore, inspired by \citet{dad}, we argue that the key to NAR text generation is to enhance the learning of token dependency---each token should be generated depending on forward and backward generated tokens.


In this paper, we propose \textsc{\textbf{Elmer}}: an \textbf{\underline{E}}fficient and Effective P\textbf{\underline{LM}} for NAR t\textbf{\underline{E}}xt gene\textbf{\underline{R}}ation, to \textit{explicitly} learn the bi-directional token dependency. Typically, most NAR models predict tokens simultaneously only at the last layer, thus making the token prediction unaware of tokens generated in other positions. To address this issue, we propose to generate tokens at different layers and the upper-layer token generation can depend on lower-layer generated tokens from both left and right. In this way, our model can explicitly learn the dependency between tokens from different layers while enjoying full parallelism in NAR decoding, as shown in Figure~\ref{fig-model}. 
To this end, we propose to extend the early exit technique~\cite{LiSSYQH20} to NAR text generation: if there is sufficient confidence to generate a token at a lower layer, the model is allowed to exit at this layer and make the prediction without passing through the upper layers.

Furthermore, instead of exiting at a fixed layer for a token, we aim to predict each token at different layers for learning diverse token dependencies in NAR text generation. Thus, inspired by XLNet \cite{xlnet}, we further propose a novel pre-training objective based on early exit, \ie Layer Permutation Language Modeling (LPLM), to help \textsc{Elmer} learn complex token dependencies. Given a sequence, LPLM will permute the exit layer (from 1 to the maximum layer) for each token and maximize the NAR text generation probability \textit{w.r.t.} all possible exit layer permutations of the sequence. Through LPLM, each token is able to exit at different layers and attend to all other tokens from both forward and backward positions. In this way, LPLM could effectively capture diverse token dependencies from large-scale corpora. Pre-trained with the general LPLM, \textsc{Elmer} can adapt to downstream text generation tasks and datasets by using specific early exit strategies.

To the best of our knowledge, we are the first to introduce the idea of early exit to NAR text generation. We fine-tune \textsc{Elmer} on three popular text generation tasks. Experiments show that \textsc{Elmer} significantly improves the best NAR models by +4.71 ROUGE-1 on XSUM, +1.79 METEOR on SQuAD v1.1, and +2.26 Distinct-2 on PersonaChat, and narrows the performance gap with auto-regressive PLMs (\eg \textsc{Elmer} (29.92) vs BART (30.61) ROUGE-L on XSUM) while achieving over 10x faster inference.

\section{Related Work}

\paratitle{Pre-trained Language Models.} Recent years have
witnessed remarkable achievement of PLMs in text generation
tasks~\cite{LiTZW21}. Most PLMs adopt an AR paradigm to generate texts during pre-training and fine-tuning. The work based on GPT~\cite{gpt2,gpt3} converts different tasks into language modeling by sequentially predicting tokens. BART~\cite{bart} employs an auto-regressive decoder to recover the corrupted text in pre-training. T5~\cite{t5} masks word spans from input texts and then sequentially predicts masked tokens. \citet{mvp} pre-trains a text generation model using labeled datasets with multi-task learning. \citet{LiTNWZ22} leverages prompts to effectively fine-tune text generation models. Differently, our PLM, \textsc{Elmer}, adopts a NAR schema to generate texts, which leads to a very low latency in inference. 

 
\paratitle{Non-autoregressive Text Generation.} Recently, there is a wide range of studies for NAR text generation~\cite{Gu0XLS18,GhazvininejadLL19,bang}. Among them, \citet{Gu0XLS18} is the first to propose NAR paradigm to reduce the inference latency. 
\citet{GhazvininejadLL19} iteratively masks and predicts a fraction of tokens that the model is least confident about. 
Several groups aim to learn  accurate input-output mapping. For example, \citet{SahariaCSN20} and \citet{LibovickyH18} use connectionist temporal classification to perform latent alignment in NAR models. Our work is closely related to BANG~\cite{bang}, a PLM bridging the NAR and AR generation. We differ in that we use early exit to predict tokens at different layers, which can help NAR models learn the forward and backward token dependency. Moreover, we propose a novel pre-training objective based on early exit, LPLM, for learning diverse token dependencies by permuting the exit layer for each token.

\section{Preliminaries}
\label{sec-pre}

Generally, the goal of text generation is to model the conditional probability $\text{Pr}(\mathcal{Y}|\mathcal{X})$, where $\mathcal{X}=\langle x_1,\dots,x_n \rangle$ and $\mathcal{Y}=\langle y_1,\dots,y_m \rangle$ denote the input text and output text respectively and each consists of a sequence of tokens from a vocabulary $\mathcal{V}$. There are three common generation paradigms to model the conditional probability $\text{Pr}(\mathcal{Y}|\mathcal{X})$, \ie autoregressive (AR), non-autoregressive (NAR), and semi-nonautoregressive (Semi-NAR) generation.

\paratitle{AR Generation.} AR generation models predict the output text based on a left-to-right factorization as:
\begin{equation}
    \text{Pr}(\mathcal{Y}|\mathcal{X}) = \prod_{t=1}^{m} \text{Pr}(y_t|y_{\textless t}, \mathcal{X}),
\end{equation}
where each token $y_t$ is generated based on the input text $\mathcal{X}$ and previous tokens $y_{\textless t}$. Note that AR generation only models the forward token dependency. The token-by-token fashion makes AR generation process hard to be parallelized. Most of existing text generation PLMs adopt AR approach~\cite{gpt2,bart,t5}.

\paratitle{NAR Generation.} In contrast to AR models, NAR text generation models predict each token in output text simultaneously as follows, without modeling the forward or backward token dependency:
\begin{equation}
    \text{Pr}(\mathcal{Y}|\mathcal{X}) = \prod_{t=1}^{m} \text{Pr}(y_t|\mathcal{X}), \label{eq-nar}
\end{equation}
where each token $y_t$ is predicted only based on the input text $\mathcal{X}$. The independence assumption makes NAR generation process  parallelizable, thus significantly accelerating the inference speed~\cite{Gu0XLS18}. While, in the absence of token dependency, 
the generation quality of NAR models is lower than their AR counterparts~\cite{WangTHQZL19}.

\paratitle{Semi-NAR Generation.} Semi-NAR generation is formalized between AR and NAR generation as:
\begin{equation}
    \text{Pr}(\mathcal{Y}|\mathcal{X}) = \prod_{t=1}^{m} \text{Pr}(y_t|\mathcal{Y}_{c_t}, \mathcal{X}),
\end{equation}
where each token $y_t$ is conditioned on the input text $\mathcal{X}$ and a visible part $\mathcal{Y}_{c_t}$ of the output text $\mathcal{Y}$. $\mathcal{Y}_{c_t}$ is designed differently to balance inference latency and accuracy~\cite{SternCKU19,LeeMC18}.
Note that the lower-layer generated tokens in our model is similar to the visible part $\mathcal{Y}_{c_t}$. While, our model keeps the advantage of full parallelism in contrast to iterative Semi-NAR methods.

In this paper, we mainly focus on the NAR approach, considering both effectiveness and efficiency for text generation models.


\section{Approach}

\begin{figure}[tb]
	\centering
	\includegraphics[width=0.47\textwidth]{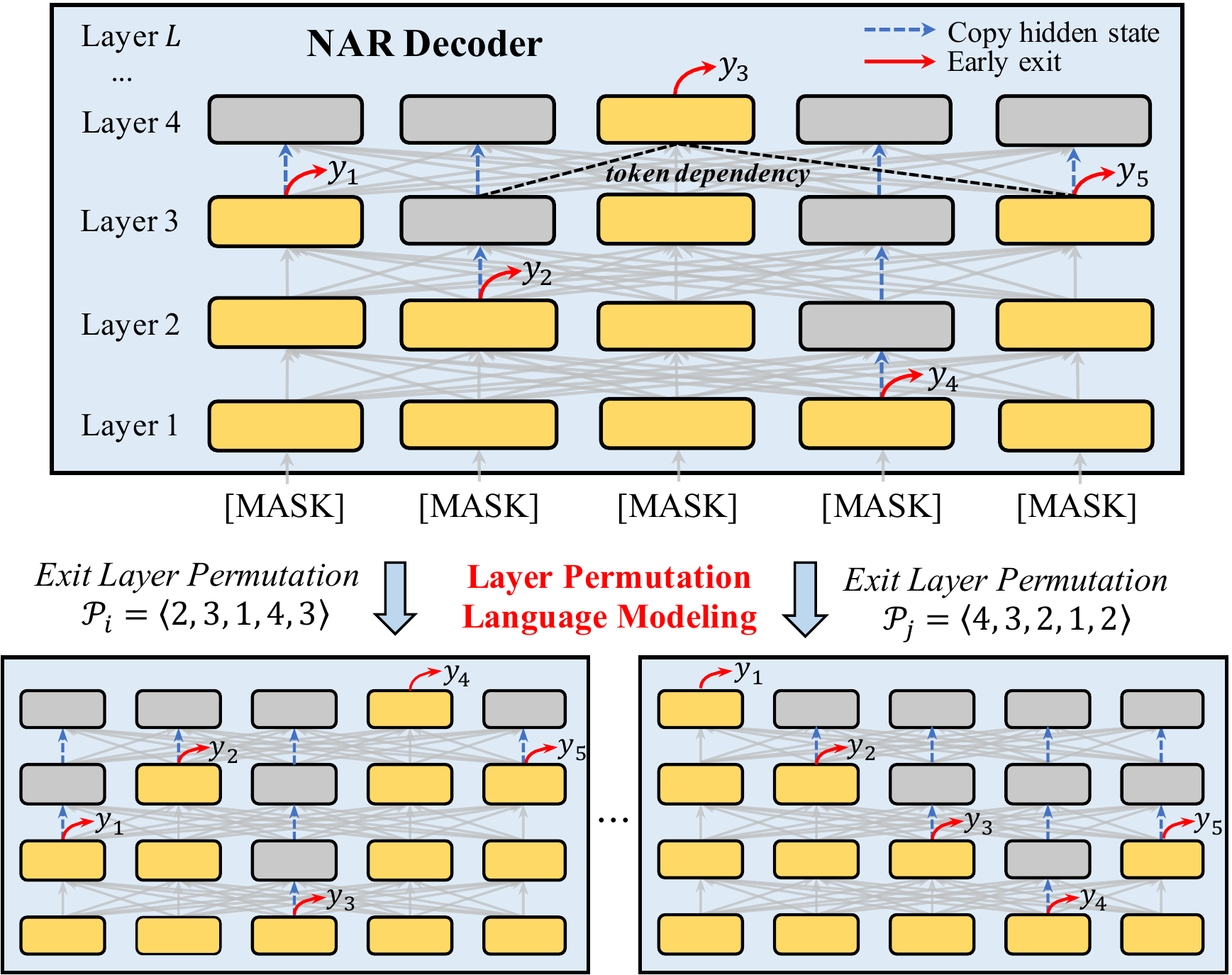}
	\caption{Overview of our proposed model \textsc{Elmer}.}
	\label{fig-model}
	\vspace{-0.2cm}
\end{figure}

Our proposed NAR text generation PLM, \textsc{Elmer}, is depicted in Figure~\ref{fig-model}. \textsc{Elmer} aims to enhance the modeling of token dependency for NAR models.
With early exit, tokens exiting at different layers can build the bi-directional token dependency with each other. Moreover, we design Layer Permutation Language Modeling (LPLM) to pre-train \textsc{Elmer} by permuting the exit layer for each token. Next, we will describe each part in detail.

\subsection{Early Exit for Dependency Modeling}
\label{}


Most NAR models simultaneously predict tokens only at the last layer~\cite{mist,dad}, which makes the current token generation unaware of the tokens generated in other positions. Thus, to model the bi-directional token dependency (both forward and backward), we propose to predict tokens at different layers by leveraging the early exit technique~\cite{LiSSYQH20}. In this way, the upper-layer token generation can depend on  tokens generated at lower layers from both left and right.

\paratitle{NAR Transformer.}  \textsc{Elmer} is built on the Transformer encoder-decoder architecture~\cite{VaswaniSPUJGKP17}. Both encoder and decoder consist of $L$ stacked layers where each layer contains several sub-layers (\eg multi-head self-attention and feed-forward network). Unlike the original Transformer decoder that auto-regressively generates text, our model uses NAR fashion to generate tokens simultaneously. Given a pair of input-output text $\langle \mathcal{X}, \mathcal{Y} \rangle$, $\mathcal{X}$ is fed into the encoder and processed as hidden states $\bm{S}=\langle \bm{s}_1,...,\bm{s}_n \rangle$. We then feed a sequence of ``\texttt{[MASK]}'' tokens into the NAR decoder to generate every token in output text $\mathcal{Y}$ in parallel.

Specifically, we first replace the original masked multi-head attention in decoder with bi-directional multi-head attention akin to the encoder. For each ``\texttt{[MASK]}'' token in the $t$-th position, the $L$ decoder layers process it to hidden states $\{\bm{h}^l_t\}_{1\leq l \leq L}$ as:
\begin{eqnarray}
    \bm{h}^l_t &=& \text{Layer}^l (\bm{h}^{l-1}_{1\leq t \leq T}, \bm{S}), \\
    \bm{h}^0_t &=& \text{Embed}(\texttt{[MASK]}),
\end{eqnarray}
where $\text{Layer}^l(\cdot)$ denotes the $l$-th layer, $\text{Embed}(\cdot)$ is the sum of word embedding and position embedding, and $T$ is the maximum length of the decoder. The output distribution for predicting the $t$-th token $y_t$ is computed by feeding the last decoder state $\bm{h}^L_t$ into a softmax classifier (parameterized by $\bm{W}_c$) as:
\begin{equation}
    \text{Pr}(y_t|\bm{h}^L_t) = \text{softmax}(\bm{W}_c \bm{h}^L_t).
\end{equation}
Prior NAR models require to determine the output length, thus an extra module for length prediction is always needed. Instead, following \citet{SuCWVBLC21}, we let \textsc{Elmer} dynamically adjust the output length by emitting an end token (\ie \texttt{[EOS]}) at any position. After the entire generation is completed, the final sequence ranges from the beginning to the position where the first end token is emitted.

\paratitle{Early Exit Off-Ramps.} 
Instead of predicting all tokens simultaneously at the last layer, we propose to generate tokens at different layers for learning the bi-directional token dependency. Hence, by leveraging early exit~\cite{LiSSYQH20}, if a token is  predicted with sufficient confidence at a lower layer, the model is allowed to exit and predict without passing through the upper layers. In this way, the upper-layer tokens can be generated depending on forward and backward tokens generated at the lower layers. 
The bi-directional token dependency can further alleviate the hallucination issue in NAR generation~\cite{Gu0XLS18}, where the generated texts tend to be ungrammatical with repetitions. 

Specifically, we inject the ``\emph{off-ramps}''~\cite{LiSSYQH20}, which make predictions with intermediate hidden states, at each decoder layer. The off-ramp can be simply implemented by a softmax classifier. For an off-ramp at the $l$-th layer, it makes the token prediction as:
\begin{eqnarray}
    \text{Pr}(y_t|\bm{h}^l_t) &=& \text{Off-Ramp}^l(\bm{h}^l_t), \\ 
    &=& \text{softmax}(\bm{W}_c^l\bm{h}^l_t), \label{off-ramp}
\end{eqnarray}
where the $l$-th off-ramp is parameterized by $\bm{W}_c^l$. These off-ramps can be specified independently or share the weights across $L$ layers. Different from previous early-exit work that makes sentence-level prediction~\cite{XinTLYL20,LiaoZRSSH21}, our early exit is built at token level.

During training, if a token has been predicted at the $l$-th  layer early, the hidden state $\bm{h}^l_t$ will not be updated in upper layers. Thus, in our model, we directly copy the last hidden state $\bm{h}^l_t$ (exit at layer $l$) to the subsequent layers following~\cite{ElbayadGGA20,LiSSYQH20}. Since the last hidden state predicts tokens with sufficient confidence, it has contained the predicted token information provided for the upper-layer tokens generation. 


\subsection{Layer Permutation Pre-training}
\label{sec-pretraining}

The NAR models equipped with early exit predict each token at a fixed layer. To learn diverse token dependencies, we design a novel pre-training objective based on early exit, Layer Permutation Language Modeling (LPLM), where each token can exit at different layers. 
This is different from most prior work that focuses on designing small-scale NAR models for specific tasks such as translation. In contrast, we pre-train a general large-scale PLM on massive corpora following~\citet{bang}. Our PLM can adapt to various downstream tasks.

\paratitle{Layer Permutation Language Modeling.} Permutation language modeling was first proposed in XLNet~\cite{xlnet} by permuting the factorization order. For a sequence with length $T$, there are $T!$ permutation orders to consider in autoregressive factorization. In our LPLM, rather than performing permutation on the sequence length, we permute the exit layer for each token. In particular, for each token in a sequence, we assume that it can exit at any of $L$ layers. Therefore, there are $L^T$ exit layer permutations for a sequence. Intuitively, if model parameters are shared across all exit layer permutations, each token can build diverse semantic dependencies with tokens in all positions.

Formally, let $\mathcal{P}_{\mathcal{Y}}=\{\bm{p}:\langle l_1,...,l_t,...,l_T \rangle \}$ be the set of all possible exit layer permutations of the sequence $\mathcal{Y}$ with length $T$. We use $l_t$ ($1\leq l_t \leq L$) to denote the exit layer of the $t$-th token. Then, for a permutation $\bm{p} \in \mathcal{P}_{\mathcal{Y}}$, the NAR generation probability (Eq.~\ref{eq-nar}) based on LPLM can be expressed as:
\begin{equation}\label{lp-prob}
    \text{Pr}(\mathcal{Y}|\mathcal{X}) = \prod_{t=1}^{T} \text{Pr}(y_t|\mathcal{X}) = \prod_{t=1}^{T} \text{Pr}(y_t|\bm{h}^{l_t}_{t}), 
\end{equation}
where the model exits at the $l_t$-th decoder layer and predicts the $t$-th token $y_t$ with the hidden state $\bm{h}^{l_t}_{t}$. $\text{Pr}(y_t|\bm{h}^{l_t}_{t})$ can be computed using Eq.~\ref{off-ramp}.

During pre-training, for a sequence in our corpora, we sample $k$ layer permutations at each time and compute the output probability with Eq.~\ref{lp-prob}. 
With the layer permutation operation, each token can exit at different layers, thus LPLM can effectively learn diverse token dependencies from large-scale corpora.
Moreover, typical early exit methods usually need to set thresholds to estimate the exit layer~\cite{ElbayadGGA20,LiSSYQH20}, which is not flexible for large-scale pre-training.
Our proposed layer permutation naturally avoids to make exit estimations in large-scale pre-training.

\paratitle{Pre-training Tasks.} Following BART~\cite{bart}, our model is trained by first feeding the corrupted text to encoder and then reconstructing the original text by decoder in a NAR manner using the above LPLM. We mainly adopt two useful document corruption methods:

\textbullet~\emph{Sentence Shuffling:} The original text is first divided into sentences according to full stops, and then these sentences are randomly shuffled.

\textbullet~\emph{Text Infilling:} Based on the shuffled text, we sample $15\%$ spans with lengths drawn from a Poisson distribution ($\lambda=3$). Following BART \cite{bart}, each span is replaced with a single ``\texttt{[MASK]}'' token and the model can learn how many tokens in a span should be predicted.

\subsection{Downstream Fine-Tuning}
\label{sec-fine-tune}

Our pre-trained model can be fine-tuned for various downstream text generation tasks. In fine-tuning phase, it becomes possible to precisely estimate the exit layer for each token with small-scale and task-specific datasets. Here, we mainly consider two early exit methods, \ie hard and soft early exit.

\paratitle{Hard Early Exit.} The most straightforward way is to calculate the exit confidence and set a threshold. Following prior work~\cite{XinTLYL20}, we quantify the exit confidence for token prediction using the \textit{entropy} of the output probability distribution:
\begin{equation}
    H(y_t)_{entropy} =-\sum \text{Pr}(y_t|\bm{h}^l_t) \cdot \log \text{Pr}(y_t|\bm{h}^l_t),
\end{equation}
where $\text{Pr}(y_t|\bm{h}^l_t)$ is computed as Eq.~\ref{off-ramp}. In this way, a low entropy means a high confidence. Specifically, at the $l$-th off-ramp, our \textsc{Elmer} model 
will compute the entropy of its output distribution $H(y_t)$ and then compare with a pre-defined threshold $\delta$ to determine whether the model should exit here or continue to the next layer.

\paratitle{Soft Early Exit.} The hard early exit method only makes one prediction for each token. Following prior work~\cite{abs-2110-07515}, the soft variant predicts the outputs at every decoder layer and the prediction is fed into the next layer for further improvement. In particular, at the $l$-th layer for the $t$-th position, we predict the most probable word $\hat{y}^l_t$ using the $l$-th off-ramp (Eq.~\ref{off-ramp}) as follows:
\begin{equation}
    \hat{y}^l_t = \arg \max \text{Pr} (y^l_t | \bm{h}^l_t).
\end{equation}
Then, we concatenate the word embedding of $\hat{y}^l_t$ with the current hidden state $\bm{h}^l_t$ and process it by a linear layer as follows:
\begin{equation}
    \tilde{\bm{h}}^l_t = \bm{W} [\text{Embed}(\hat{y}^l_t); \bm{h}^l_t],
\end{equation}
where $\bm{W}$ is a learnable matrix and $\tilde{\bm{h}}^l_t$ is the updated hidden state of the $l$-th layer, which will be taken as input to the next $l+1$-th layer. Compared with the hard early exit, the soft early exit is able to calibrate the token prediction at each layer.

\subsection{Time Complexity Analysis}

Since AR and NAR models adopt the same encoder architecture, the difference in time complexity is dominated by decoders. To generate a sequence with length $T$, an AR decoder with $L$ layers has a time complexity $\mathcal{O}(LT^2)$ quadratic in sequence length. In contrast, a NAR decoder has a linear time complexity $\mathcal{O}(LT)$. This is because the attention computation in NAR decoders can be parallelized across all positions. While,
our \textsc{Elmer} model generates tokens at different layers. Let  $\overline{L} (<L)$ denote the average exit layer for a sequence, the time complexity will further be decreased to $\mathcal{O}(\overline{L}T)$.

\section{Experiments}

In this section, we detail the experimental setup and then highlight the main takeaways of our results.

\subsection{Experimental Setup}
\paratitle{Pre-training Setup.} We pre-train \textsc{Elmer} based on the 16GB corpus (including English Wikipedia and BookCorpus). For our model, we use 6 layers in both encoder and decoder, with a dimension of 768 in hidden states, consistent with the base version of many AR and NAR pre-trained models~\cite{bart,prophetnet,bang}. We pre-train our model from scratch with a learning rate of 2e-4 and a minibatch size of 4096. We adopt the dictionary from BART~\cite{bart}. In pre-training, we share the off-ramp weights across all layers and sample 10 exit layer permutations for a sequence.

\paratitle{Fine-Tuning Datasets.} Following prior work~\cite{bang}, we fine-tune \textsc{Elmer} on three text generation tasks and datasets: (1) \textbf{XSUM}~\cite{xsum} is a news summarization dataset containing 227K news article and single-sentence summary pairs; (2) \textbf{SQuAD v1.1}~\cite{squad} is a question generation dataset, containing 98K triples of passage, question, and answer. We concatenate the passage and answer as input and predict the question; (3) \textbf{PersonaChat}~\cite{personachat} is a dialog generation dataset, containing 150K triples of persona profile, dialogue history, and response. We   concatenate the profile and dialogue history as input and generate the response. The statistics of datasets are shown in Appendix~\ref{app-dataset}. 


\begin{table*}[t]
	\small
	\centering
	\begin{tabular}{l l  r r r r r r r r}
		\toprule
		\multirow{2.5}{*}{\textbf{Type}} & \multirow{2.5}{*}{\textbf{Models}} & \multicolumn{3}{c}{\textbf{XSUM}} & \multirow{2.5}{*}{\tabincell{c}{Latency$\downarrow$\\(ms/Sample)}} & \multicolumn{3}{c}{\textbf{SQuAD v1.1}} & \multirow{2.5}{*}{\tabincell{c}{Latency$\downarrow$\\(ms/Sample)}}\\ 
		\cmidrule{3-5}\cmidrule{7-9}
		&  & R-1$\uparrow$  & R-2$\uparrow$ & R-L$\uparrow$ & & R-L$\uparrow$  & B-4$\uparrow$ & MT$\uparrow$ \\ 
		\midrule
		\multirow{4}{*}{\textbf{AR}} & \textbf{Transformer} & 30.66 & 10.80 & 24.48 & 262.47 (22.0x) & 29.43 & 4.61 & 9.86	& 159.49  (13.4x) \\
		& \textbf{MASS} & 39.70 & 17.24 & 31.91 & 196.17 (16.4x) & 49.48 & 20.16 & 24.41 & 132.46 (11.1x) \\
		& \textbf{BART} & 38.79 & 16.16 & 30.61 &  185.19 (15.5x) & 42.55 & 17.08 & 23.19  & 114.00 (\;\,9.6x)  \\ 
		& \textbf{ProphetNet} & 39.89 & 17.12 & 32.07 & 817.80 (68.5x) & 48.00 & 19.58 & 23.94  & 456.51 (38.4x) \\ 
		\cmidrule{1-10}
		\multirow{5}{*}{\textbf{Semi-NAR}} & \textbf{InsT} & 17.65 & 5.18 & 16.05 & 63.37 (5.3x) & 29.98 & 2.34 & 8.15  & 67.61 (5.7x) \\
		& \textbf{iNAT} & 26.95 & 6.88 & 22.43 & 31.27 (2.6x) & 32.34 & 3.16 & 9.18  & 31.59 (2.7x) \\
		& \textbf{CMLM} & 29.12 & 7.70 & 23.04 & 113.64 (9.5x) & 29.60 & 3.89 & 9.70  & 106.84 (9.0x) \\
		& \textbf{LevT} & 25.33 & 7.40 & 21.48 & 101.01 (8.5x) & 30.81 & 2.68 & 9.40  & 116.41 (9.8x) \\
		& \textbf{BANG} & 34.71 & 11.71 & 29.16 & 109.77 (9.2x) & 47.39 & 17.62 & 21.69  & 111.11 (9.3x) \\
		\cmidrule{1-10}
		\multirow{7.5}{*}{\textbf{NAR}} & \textbf{NAT} & 24.04 & 3.88 & 20.32 & 17.47 (1.5x) & 31.51 & 2.46 & 8.86  & 17.11 (1.4x)\\
		& \textbf{iNAT} & 24.02 & 3.99 & 20.36 & 16.94 (1.4x) & 32.44 & 2.33 & 8.84  & 16.52 (1.4x)\\
		& \textbf{CMLM} & 23.82 & 3.60 & 20.15 & 16.88 (1.4x) & 31.58 & 2.51 & 8.85  & 16.41 (1.4x)\\
		& \textbf{LevT} & 24.75 & 4.18 & 20.87 & 27.72 (2.3x) & 31.38 & 2.27 & 9.14  & 27.52 (2.3x)\\
		& \textbf{BANG} & 32.59 & 8.98 & \underline{27.41} & 15.97 (1.3x) & \textbf{44.07} & \underline{12.75} & \underline{18.99}  & 15.69 (1.3x)\\
		\cmidrule{2-10}
		& \textbf{\textsc{Elmer}-Hard} & \underline{34.54} & \underline{9.78} & 26.08 & \underline{12.39} (1.0x) & 37.94 & 11.77 & 18.01 & \underline{12.24} (1.0x) \\
		& \textbf{\textsc{Elmer}-Soft} & \textbf{38.30} & \textbf{14.17} & \textbf{29.92} & \textbf{11.94} (1.0x) & \underline{40.22} & \textbf{13.49} & \textbf{20.08} & \textbf{11.90} (1.0x) \\
		\bottomrule   
	\end{tabular}
	\caption{A comparison between \textsc{Elmer} and baselines on XSUM and SQuAD v1.1 datasets. R-1/2/L, B-4, and MT are short-hands for ROUGE-1/2/L, BLEU-4, and METEOR. \textbf{\textsc{Elmer}-Hard} and \textbf{\textsc{Elmer}-Soft} denote fine-tuning \textsc{Elmer} with hard and soft early exit strategies, respectively. \textbf{Bold} and \underline{underline} fonts denote the best and second best methods within NAR models. Our baseline results are collected from \cite{LiuYGQZJCFSGWCJ21} and \cite{bang}.}
	\label{tab:xsum-squad}
\end{table*}

\paratitle{Baselines}. We compare \textbf{\textsc{Elmer}} to existing popular  AR, NAR, and Semi-NAR generation models.

For AR generation, we experiment with a vanilla Transformer model and three PLMs:

\textbullet~\textbf{Transformer}~\cite{VaswaniSPUJGKP17}. It is an AR generation model without pre-training. To date, Transformer has become the backbone of many text generation PLMs and our \textsc{Elmer} model.

\textbullet~\textbf{MASS}~\cite{mass}, \textbf{BART}~\cite{bart}, and \textbf{ProphetNet}~\cite{prophetnet}. These are three representative PLMs for AR text generation, whose pre-training objectives vary from denoising text to future $n$-gram prediction. For a fair comparison, we adopt the base version of these PLMs.

For NAR and Semi-NAR generation, we evaluate six models with varying decoding strategies:

\textbullet~\textbf{NAT}~\cite{Gu0XLS18}. It is the first proposed NAR text generation model. This baseline adds a module in the encoder to predict fertilities, acting as a global plan for the parallel generation.

\textbullet~\textbf{InsT}~\cite{SternCKU19}. It is a Semi-NAR text generation model leveraging insertion operations. It repeatedly inserts tokens at multiple locations based on the partially inserted sequence.

\textbullet~\textbf{iNAT}~\cite{LeeMC18}, \textbf{CMLM}~\cite{GhazvininejadLL19}, \textbf{LevT}~\cite{GuWZ19}, and \textbf{BANG}~\cite{bang}. These four baselines are both NAR and Semi-NAR text generation models. Among them, BANG is the recent state-of-the-art PLM for NAR text generation. 


\begin{table*}[t]
	\small
	\centering
	\begin{tabular}{l l  r r r r r r}
		\toprule
		\multirow{2.5}{*}{\textbf{Type}} & \multirow{2.5}{*}{\textbf{Models}} & \multicolumn{5}{c}{\textbf{PersonaChat}} & \multirow{2.5}{*}{\tabincell{c}{Latency$\downarrow$\\(ms/Sample)}} \\ 
		\cmidrule{3-7}
		&  & BLEU-1$\uparrow$  & BLEU-2$\uparrow$ & Distinct-1$\uparrow$ & Distinct-2$\uparrow$ & Overall$\uparrow$ & \\ 
		\midrule
		\multirow{4}{*}{\textbf{AR}} & \textbf{Transformer} & 41.56 & 32.95 & 0.30 & 0.80 & 18.90 & 138.31  (11.9x) \\
		& \textbf{MASS} & 41.06 & 35.75 & 1.40 & 6.90 & 21.28 &  112.13 (\;\,9.7x) \\
		& \textbf{BART} & 47.60 & 39.36 & 1.10 & 6.10 & 23.54 &  106.15 (\;\,9.2x) \\ 
		& \textbf{ProphetNet} & 46.00 & 38.40 & 1.30 & 7.30 & 23.25 & 392.79 (33.9x) \\ 
		\cmidrule{1-8}
		\multirow{5}{*}{\textbf{Semi-NAR}} & \textbf{InsT} & 12.63 & 9.43 & 0.10 & 0.30 & 5.62 & 65.27 (5.6x) \\
		& \textbf{iNAT} & 41.17 & 32.13 & 0.10 & 1.10 & 18.63 & 43.25 (3.7x) \\
		& \textbf{CMLM} & 44.38 & 35.18 & 0.10 & 0.80 & 20.12 & 105.82 (9.1x) \\
		& \textbf{LevT} & 24.89 & 18.94 & 0.10 & 0.60 & 11.13 & 80.26 (6.9x) \\
		& \textbf{BANG} & 39.82 & 30.72 & 1.90 & 14.20  & 21.66 & 109.17 (9.4x) \\
		\cmidrule{1-8}
		\multirow{7.5}{*}{\textbf{NAR}} & \textbf{NAT} & \textbf{31.53} & \textbf{24.17} & 0.10 & 0.80 & 14.15 & 17.86 (1.5x) \\
		& \textbf{iNAT} & 30.56 & 23.38 & 0.10 & 0.70  & 13.69 & 16.40 (1.4x) \\
		& \textbf{CMLM} & 31.44 & \underline{24.06} & 0.10 & 0.60  & 14.05 & 16.26 (1.4x) \\
		& \textbf{LevT} & 26.92 & 20.47 & 0.00 & 0.40  & 11.95 & 27.56 (2.4x) \\
		& \textbf{BANG} & 31.11 & 23.90 & \underline{2.50} & \underline{22.70} & \underline{20.05} & 14.89 (1.3x) \\
		\cmidrule{2-8}
		& \textbf{\textsc{Elmer}-Hard} & 29.43 & 21.89 & 1.56 & 21.45 & 18.58 &  \underline{12.01} (1.0x) \\
		& \textbf{\textsc{Elmer}-Soft} & \underline{31.45} & 23.99 & \textbf{3.66} & \textbf{24.96} & \textbf{21.02} &  \textbf{11.59} (1.0x) \\
		\bottomrule   
	\end{tabular}
	\caption{A comparison between \textsc{Elmer} and baselines on PersonaChat with respect to automatic evaluation metrics.}
	\label{tab:personachat}
\end{table*}

\paratitle{Evaluation Metrics}. To evaluate the \textit{effectiveness} of different models, we adopt four evaluation metrics: ROUGE-$n$ assesses the text quality by computing the overlapping $n$-grams between the generated and real texts~\cite{lin2004rouge}; BLEU-$n$ computes the co-occurrence ratio of $n$-grams between the generated and real texts~\cite{papineni2002bleu}; METEOR assesses word-to-word matches between the generated and real texts based on the harmonic mean of the unigram precision and recall~\cite{BanerjeeL05}; and Distinct-$n$ measures the diversity degree by calculating the number of distinct $n$-grams in generated texts~\cite{LiGBGD16}. To examine the \textit{efficiency}, we set the batch size as 1 at inference to calculate the per-sample inference latency under the same parameter setting following~\cite{bang}. More details about our experiments can be found in Appendix~\ref{app-configuration}.

\subsection{Main Results}
\label{sec-full}

Table~\ref{tab:xsum-squad} and Table~\ref{tab:personachat} display the results of \textsc{Elmer} and baselines on three text generation datasets.

First, \textsc{Elmer}-Soft outperforms NAR and Semi-NAR baselines on almost all datasets and metrics. Compared to the best NAR BANG, \textsc{Elmer}-Soft achieves prominent gains in effectiveness metrics such as +5.71 ROUGE-1 in XSUM, +1.09 METEOR in SQuAD v1.1, and +2.26 Distinct-2 in PersonaChat. 
These considerable gains clearly demonstrate the \textit{effectiveness} of our \textsc{Elmer} model. In contrast to these NAR and Semi-NAR models, our model leverages the early exit technique to explicitly model the forward and backward token dependency during parallel NAR decoding.

Second, \textsc{Elmer} consistently achieves comparable or better results than the AR baseline without pre-training, \ie Transformer, and further narrows the performance gap between NAR and AR PLMs. 
For example, the performance gap between BANG and BART in XSUM and SQuAD is 3.20 ROUGE-L and 4.20 METEOR, which has now been decreased by \textsc{Elmer}-Soft to 0.69 and 3.11.  \textsc{Elmer}-Soft also obtains the best Distinct scores in PersonaChat. 
Diversity is critical for dialogue generation to avoid boring or useless responses. Besides, the lower BLEU scores than NAT probably because the diverse generated texts by \textsc{Elmer}-Soft makes a reasonable difference from real texts. For those models without pre-training such as CMLM and NAT, they tend to generate highly frequent words and common phrases such as ``\textit{I, an, the}'', which share much overlapped and repetitive parts with the target output. By contrast, our pre-trained model generates diverse responses, which is more critical to avoid boring or useless responses in dialogue generation. 

Finally, in terms of \textit{efficiency}, \textsc{Elmer} achieves much faster inference speed than all NAR and AR models, \textit{w.r.t.} the per sample inference latency. For example, the inference latency of AR Transformer model in XSUM, SQuAD v1.1, and PersonaChat are 262.47ms, 159.49ms, and 138.31ms per sample, while those of \textsc{Elmer}-Soft are 11.94ms, 11.90ms, and 11.59ms, which amount to 22.0, 13.4, and 11.9 times speedup. Compared to AR PLMs, we achieve over 10x inference speedup, especially the highest 68.5x speedup in XSUM. Moreover, \textsc{Elmer}-Soft is slightly faster than other NAR baselines. We speculate that this speedup is mainly due to the fact that prior NAR models require to predict the output length such as NAT or rely on multiple insertion and deletion operations such as LevT.

\begin{table}[t]
	\small
	\centering
	\setlength\tabcolsep{3.5pt}
	\begin{tabular}{lrrr}
		\toprule
		 \multirow{2.5}*{\textbf{Models}} & \multicolumn{3}{c}{\textbf{XSUM}} \\
		 \cmidrule{2-4}
		 & \textbf{ROUGE-1} & \textbf{ROUGE-2} & \textbf{ROUGE-L} \\
		\midrule
		NAR LevT    & 24.75 &  4.18  &  20.87 \\
		\midrule
		\textsc{Elmer}-Hard       & 34.54     & 9.78     & 26.08  \\
		-w/o pre-training & 28.56  & 5.33 & 21.96 \\
		\textsc{Elmer}-Soft       & 38.30     & 14.17     & 29.92   \\
		-w/o pre-training & 30.45  & 7.37 & 24.00 \\
		\bottomrule
	\end{tabular}
	\caption{Ablation study on XSUM dataset.}
	\label{tab:ablation}
\end{table}

\subsection{Detailed Analysis} 

We report detailed analysis of our model on XSUM dataset -- we have similar findings on other datasets. 


\paratitle{Ablation Study.} Our \textsc{Elmer} model is the first one to adopt the early exit technique to conduct NAR text generation. To learn more diverse token dependencies, we proposed the LPLM objective (Sec.~\ref{sec-pretraining}) based on early exit to pre-train \textsc{Elmer}. Here, we evaluate the effects of early exit and large-scale LPLM training by directly applying the hard and soft early exit methods to NAR text generation without pre-training. We compare the early-exit variants with the best non-pretrained NAR model in XSUM, \ie LevT. The results are shown in Table~\ref{tab:ablation}. We can observe that these two early-exit variants without pre-training perform better than LevT. This indicates that introducing the early exit mechanism can improve NAR text generation quality due to the explicit modeling of the target-side dependency. Moreover, our proposed large-scale pre-training based on the LPLM objective improves NAR results significantly and consistently in both early exit strategies. By utilizing the general LPLM, our model can capture diverse token dependencies from large-scale corpora and further adapt to downstream tasks with specific early exit methods.

\begin{figure}[t]
	\centering
	\subfigure[$\delta=0.5$]{
		\centering
		\includegraphics[width=0.22\textwidth]{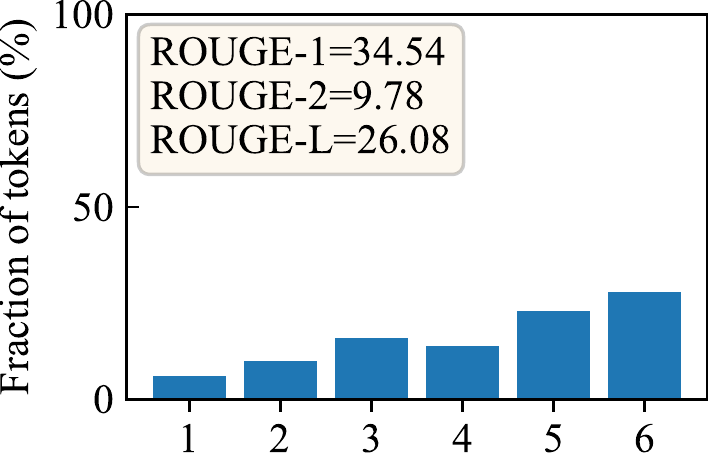}
	}
	\subfigure[$\delta=1.0$]{
		\centering
		\includegraphics[width=0.22\textwidth]{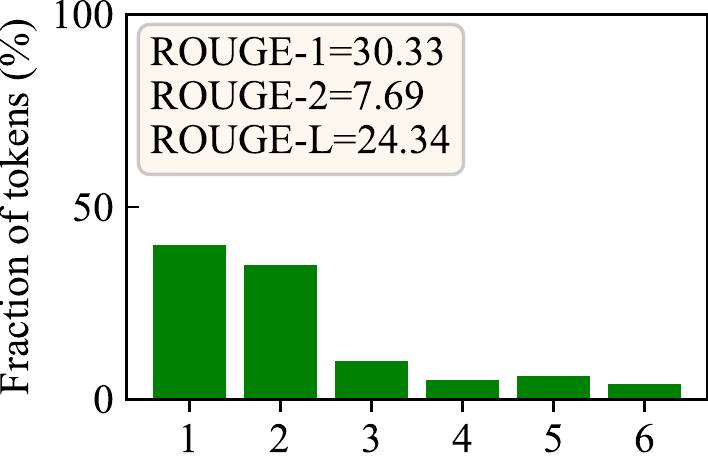}
	}
	\centering
	\caption{Varying $\delta$ for \textsc{Elmer}-Hard on XSUM.}
	\label{fig:window-size}
	\vspace{-0.2cm}
\end{figure}

\paratitle{Parameter Sensitivity Analysis.} In \textsc{Elmer}-Hard, the pre-defined entropy threshold $\delta$ is critical to determine at which layer our model will exit. 
Here, we further examine the model performance and the fraction of tokens exiting at every layers ($1\sim6$) by varying the exit threshold in the set \{0.0, 0.25, 0.5, 0.75, 1.0\}. As the threshold $\delta$ increases gradually, more tokens exit earlier. 
From the experiment, we find that low and high thresholds lead to reverse early exit situations shown by the fraction of tokens exiting at each layer. Therefore, we only present the results for two representative thresholds (0.5 and 1.0) in Figure~\ref{fig:window-size}. 
We can observe that: (1) when $\delta=0.5$, a larger fraction of tokens are generated in higher layers ($\geq3$), which means a lower threshold needs more computation to achieve enough confidence. These upper-layer ``difficult'' tokens can depend on simple tokens such as ``\textit{I}'', ``\textit{the}'', ``\textit{have}'' generated at lower layers. (2) when $\delta=1.0$, it shows an opposite trend and the performance drops slightly. The reason might be that $1.0$ is so high that the model makes a precipitate exiting decision. Finally, according to the model performance on the validation set, we set the threshold $\delta$ as 0.5 in our experiments.

\subsection{Human Evaluation} 

Despite the effectiveness of automatic evaluation, human evaluation remains critical for text generation~\cite{eva_survey}. Since human evaluation is expensive, we only focus on comparing our best performing \textsc{Elmer}-Soft with two AR and NAR PLMs, \ie BART and BANG. 

In order to reduce the variance caused by human, three workers were asked to score texts from four aspects~\cite{LiTGYYCWZW22}, \ie \textit{fluency}, \textit{informativeness}, \textit{accuracy}, and \textit{relevance}. \textit{Fluency} evaluates whether the text is well-formed and logical to read; \textit{Informativeness} measures whether the text contains useful information; \textit{Accuracy} tests whether the text describes the given content accurately; \textit{Relevance} measures whether the text is relevant to the given context. 
These four scores are rated from 1 to 5. 
We further design a Turing test~\cite{Turing50} where a human judge is asked to detect whether the given text is generated by a human. For each method, we average the scores from three human judges and then report the average results over 500 texts. From the results in Table~\ref{tab:turing-test}, we can see that our model is better than NAR model BANG in terms of \textit{fluency} (3.99 vs 3.42) and \textit{relevance} (3.63 vs 3.38). The major reason is that our model leverages the early exit to explicitly model the token dependency, which could reduce token repetitions and improves the fluency of texts. While, our model achieves a slightly worse fluency score than the AR model BART. We speculate that BART can produce some common phrases by using the AR generation mode. 

We also present some generated examples by our model in Appendix~\ref{app-examples}. It can be observe that our model can generate some common phrases occurred in real texts such as ``\textit{surface-to-air}'', ``\textit{South China Sea}'', and ``\textit{main seminary}''. These phrases can improve the fluency of NAR generated texts. By incorporating early exit and generating tokens at different layers, our model can explicitly learn the forward and backward token dependency, which can deal with the multi-modality issue to some extent. While, as a NAR model, our model still inevitably generates some repetitive stop tokens such as ``\textit{on on on}'', which can be further improved in future work.

\begin{table}[t]
	\small
	\centering
	\begin{tabular}{lr|rrrr}
		\toprule
		 \multirow{2.5}*{\textbf{Models}} & \multicolumn{5}{c}{\textbf{XSUM}} \\
		 \cmidrule{2-6}
		 & \textbf{TT (\%)} & \textbf{Flu.} & \textbf{Info.} & \textbf{Acc.} & \textbf{Rel.} \\
		\midrule
		BART       & 51.01          & 4.11          & 3.99          & 3.64          & 3.79          \\
		BANG         & 45.34 & 3.42          & 3.45          & 3.46    & 3.38          \\
		\textsc{Elmer}-Soft       & 48.90          & 3.99          & 3.52          & 3.49 & 3.63    \\
		\cmidrule{1-6}
		Gold       & 55.67          & 4.12          & 4.29          & 3.95          & 4.05         \\
		\bottomrule
	\end{tabular}
	\caption{Turing test (TT) and human scores on XSUM. Flu., Info., Acc. and Rel. denote fluency, informativeness, accuracy and relevance respectively.}
	\label{tab:turing-test}
\end{table}

\section{Conclusion}

This paper presented an efficient and effective PLM for NAR text generation, called \textsc{Elmer}. \textsc{Elmer} introduced a token-level early exit mechanism to explicitly model the semantic dependency between target tokens during parallel decoding. Moreover, we proposed a pre-training objective, Layer Permutation Language Modeling, to pre-train \textsc{Elmer} on large-scale corpora by permuting the exit layer for each token in a sequence. Experiments on text summarization, question generation, and dialogue generation demonstrate that our model can effectively improve the NAR generation quality compared to highly competitive NAR models and further narrow the performance gap with AR models while achieving higher inference efficiency. In future work, we will investigate the efficacy of utilizing the proposed model to iteratively generate texts and consider more early exit strategies in fine-tuning.

\section{Limitations}

An important limitation of \textsc{Elmer} compared with other NAR text generation models is the need for defining an appropriate early exit strategy. In pre-training, \textsc{Elmer} utilizes a general strategy, \ie layer permutation language modeling, by permuting the exit layer for each token in a sequence. However, to apply \textsc{Elmer} to downstream specific tasks and datasets, we need to design an effective and suitable estimation method to decide at which layer the model will exit and predict tokens. Also, as a PLM, \textsc{Elmer} may present  biases learned from the pre-training corpus in the output text.

In this study, we evaluate \textsc{Elmer} on three text generation tasks and datasets, but the output length of these tasks is relatively short. We should deal with long-form text generation 
which raises more challenges to the NAR paradigm.
\section{Ethical Concerns}

Current NAR text generation techniques achieve faster inference speed but suffer from several issues like multi-modality, lack of accuracy to the input, commonsense issues etc., which makes their online real-time deployment difficult. \textsc{Elmer} is an effort at rectifying some of these issues, with a focus of modeling the semantic dependency between target tokens to improve NAR generation quality. However, compared to AR models, \textsc{Elmer} outputs continue to mix multiple candidates and repeat words at times. This should be strongly considered before any direct deployment of real-world systems. 

On the other hand, the text generation technology may be potentially misused for harmful applications. When deploying \textsc{Elmer} to online real-time platforms, the high-quality text generated by our work also makes it difficult to distinguish synthetic text from human-written text, such as fake news and stories.  It is somewhat difficult to anticipate the harmful usages of our method since they often involve repurposing our model in a totally different setting or for an unexpected purpose than we planned. To alleviate this problem, we can ask for help from some classic security risk assessment frameworks such as detecting threats and potential impacts, measuring likelihood, and determining risk as a combination of likelihood and impact~\cite{blank2011guide}.

\section*{Acknowledgement}

This work was partially supported by Beijing Natural Science Foundation under Grant No. 4222027, and Beijing Outstanding Young Scientist Program under Grant No. BJJWZYJH012019100020098, and the Outstanding Innovative Talents Cultivation Funded Programs 2021 of Renmin University of China. Xin Zhao is the corresponding author.

\section*{Reproducibility}
For reproducing and reusing our work, we release the pre-trained \textsc{Elmer} model and source codes at the link:~\url{https://github.com/RUCAIBox/ELMER}. In the future, we will integrate our model into Hugging Face~\cite{huggingface} and TextBox~\cite{textbox} libraries for easy-to-use. We hope that these open-source resources will facilitate and contribute to the advancement of related research.

\bibliography{anthology}
\bibliographystyle{acl_natbib}

\newpage
\appendix

\section*{Appendix}
We provide some experiment-related information as supplementary materials. The appendix is organized into three sections:
\begin{itemize}
	\item Statistics of each dataset are presented in Appendix~\ref{app-dataset};
	\item Training settings of baselines and our model \textsc{Elmer} are presented in Appendix~\ref{app-configuration};
	\item Generated examples by our model are presented in Appendix~\ref{app-examples}.
\end{itemize}

\section{Statistics of Datasets} \label{app-dataset}
The detailed information of these three datasets is listed in Table~\ref{tab-data}.

\begin{table}[h]
	\centering
	\small
	\begin{tabular}{l|r r r r}
		\toprule[1pt]
		\textbf{Dataset} & \#Train & \#Valid & \#Test & \#Output \\
		\midrule[0.7pt]
		\textbf{XSUM} & 204,045 & 11,332 & 11,334 & 21.1 \\
		\textbf{SQuAD v1.1} & 75,722 & 10,570 & 11,877 & 11.6 \\
		\textbf{PersonaChat}	 & 122,499 & 14,602 & 14,056 & 11.9 \\
		\bottomrule[1pt]
	\end{tabular}%
	\caption{Statistics of three datasets. \#Output denotes the average number of tokens in the output texts.}
	\label{tab-data}%
\end{table}

\section{Experimental Details} \label{app-configuration}

For AR baselines, we adopt the hyper-parameters: learning rate 2e-5, batch size 20, Adam optimizer, and the maximum input and output length of 512. We fine-tune these models on each dataset for 50 epochs and select the best model. For NAR and Semi-NAR baselines, the hyper-parameters are the same as AR baselines except the number of fine-tuning epochs, since NAR models need more epochs to converge~\cite{bang}. We fine-tune these baselines and \textsc{Elmer} for 50 epochs and save checkpoints for every epoch. We select the best checkpoint based on the performance on validation set. Following BANG~\cite{bang}, the difference between semi-NAR and NAR models lies in that: (1) we select the outputs from the first iteration as the NAR outputs, since the first iteration only depends on the input text; (2) we select the outputs from the maximum (we set as 10) iteration as the semi-NAR outputs, since each iteration depends on both the input text and the last generated texts. Most settings and results are collected from GLGE~\cite{LiuYGQZJCFSGWCJ21} and BANG~\cite{bang}. To make a fair comparison with non-pretrained baselines, we also compare \textsc{Elmer} with the best NAR non-pretrained baseline LevT in ablation analysis (Table~\ref{tab:ablation}).


\section{Case Study} \label{app-examples}

We show some qualitative examples of these three datasets in Table~\ref{tab:xsum}, Table~\ref{tab:squad}, and Table~\ref{tab:pc}.

Due to the explicit modeling of the bi-direction token dependency, our model can generate some common phrases occurred in real texts such as ``surface-to-air'', ``South China Sea'', and ``main seminary''. These phrases can improve the fluency of NAR generated texts. But, as a NAR model, our model still inevitably generates some repetitive tokens such as ``on on on''. The generated examples in PersonaChat also explain that our model generates more diverse responses, which are very different from the real responses.

\begin{table*}[b]
\small
\centering
	\begin{tabular}{l | p{0.4\textwidth} | p{0.4\textwidth}}
			\toprule[1pt]
			\textbf{News} &  Reports on Wednesday suggested more than one iguana was actually filmed, with scenes then stitched together. But the BBC has said only one animal was chased by the snakes - with other iguanas only filmed for close-ups. The scene quickly went viral when it was aired last year and later won a Bafta for must-see moment. The iguana hatchling, filmed in the Galapagos, eventually got away - much to viewers' relief. The Daily Mail claimed the episode was embroiled in a "fakery row" after producer Elizabeth White told the Media Production Show: "It wasn't the same iguana, no, and often we have to augment it with other clips. "Unfortunately lizards, snakes and iguanas aren't good at 'takes'." But the BBC defended the Sir David Attenborough-fronted programme, with a spokeswoman saying: "The BBC strongly refutes any suggestion that the award-winning iguana v snakes sequence was 'faked'. "The final iguana chase in which one iguana escapes the snakes was - unusually for natural history filming - shot using two cameras, allowing us to follow both the individual iguana and the snakes' point of view. "What was captured in the field was extraordinary animal behaviour which had never been witnessed or filmed before." She added: "As is common in natural history film-making, pick-up shots were filmed separately - for example close-ups of iguana eyes - to make the story of the sequence as clear as possible for the audience. "This is absolutely in keeping with the norms of natural history film-making - and absolutely in line with the BBC's editorial policy guidelines, and was a true representation of animal behaviour." \ignore{Other BBC nature documentaries have previously been accused of faking footage. Frozen Planet, also fronted by Sir David, showed footage of newborn polar bear cubs in a den with their mother in 2011 - but it was filmed in an animal park, rather than in the wild. The BBC denied misleading viewers, with Sir David saying it would have spoiled the atmosphere to point out where the filming had taken place adding: "It's not a falsehood and we don't keep it secret either".} & Satellite images taken on 14 February appear to show two batteries of eight missile launchers and a radar system on Woody or Yongxing Island in the Paracels. The presence of missiles would significantly increase tensions in the acrimonious South China Sea dispute. China's Foreign Minister Wang Yi said reports were a Western media invention. But Mr Wang defended "the limited and necessary self-defence facilities" on islands inhabited by Chinese personnel as "consistent with the right for self-preservation and self-protection.... under the international law". Asked about the reports, US Secretary of State John Kerry attacked China's increased "militarisation" of the contested region, saying it was a "serious concern". Taiwan's defence ministry said it had "learned of an air defence missile system deployed" by the Chinese on Woody Island. It would not say how many missiles had been deployed or when, but told the BBC they would be capable of targeting civilian and military aircraft. The commander of the US Pacific Fleet confirmed the deployment to Reuters news agency. Adm Harry Harris said such a move would be "a militarisation of the South China Sea in ways" China's President Xi Jinping had pledged not to make. Japan's Chief Cabinet Secretary Yoshihide Suga said there were "serious concerns" over China's "unilateral move to change the status quo" in the region, and "we cannot accept this fact". China has been carrying out extensive land reclamation work in the region, which it says is legal and for civilian purposes. But the work has angered other countries which also claim the territory, and there is growing concern about the implications of the area becoming militarised. \ignore{The latest images of Woody Island were captured by ImageSat International. They show a close-up of a section of beach, the shape of which resembles the northern coastline as it appears on other images, and point out two missiles batteries. Each battery is made up of four launchers and two control vehicles. Two of the the launchers appear to have been erected, says the report.} \\
			\midrule[0.5pt]
			\textbf{Summaries} & The BBC has denied claims award-winning series Planet Earth II faked a nail-biting scene showing a baby iguana being chased by racer snakes . & China has deployed surface-to-air missiles on a disputed island in the South China Sea, Taiwan says . \\
			\midrule[0.5pt]
			\textbf{\textsc{Elmer}-Hard} & The BBC has defended claims the BBC nature documentary of the iguana footage of a "faked storm " by a Davidborough. & China has confirmed that new reports of on surface-air on on on of in the South China Sea . \\
			\midrule[0.5pt]
			\textbf{\textsc{Elmer}-Soft}  & The BBC has defended claims a BBC documentary about a series of from natural drama Sir David Atten was "faked". & China is the first of of air defence missiles on a disputed disputed island of the South China Sea, according to US \\
			\bottomrule[1pt]
		\end{tabular}
	\caption{Qualitative examples on XSUM dataset.} 
	\label{tab:xsum}
\end{table*}

\begin{table*}[t]
\small
\centering
	\begin{tabular}{l | p{0.4\textwidth} | p{0.4\textwidth}}
			\toprule[1pt]
			\textbf{Paragraphs} &  Moreau Seminary [SEP] The university is the major seat of the Congregation of Holy Cross ( albeit not its official headquarters , which are in Rome ) . Its main seminary , Moreau Seminary , is located on the campus across St . Joseph lake from the Main Building . Old College , the oldest building on campus and located near the shore of St . Mary lake , houses undergraduate seminarians . Retired priests and brothers reside in Fatima House ( a former retreat center ) , Holy Cross House , as well as Columba Hall near the Grotto . The university through the Moreau Seminary has ties to theologian Frederick Buechner . While not Catholic , Buechner has praised writers from Notre Dame and Moreau Seminary created a Buechner Prize for Preaching . & eight [SEP] The College of Engineering was established in 1920 , however , early courses in civil and mechanical engineering were a part of the College of Science since the 1870s . Today the college , housed in the Fitzpatrick , Cushing , and Stinson - Remick Halls of Engineering , includes five departments of study ?~@~S aerospace and mechanical engineering , chemical and biomolecular engineering , civil engineering and geological sciences , computer science and engineering , and electrical engineering ?~@~S with eight B . S . degrees offered . Additionally , the college offers five - year dual degree programs with the Colleges of Arts and Letters and of Business awarding additional B . A . and Master of Business Administration ( MBA ) degrees , respectively . \\
			\midrule[0.5pt]
			\textbf{Questions} & What is the primary seminary of the Congregation of the Holy Cross ? & How many BS level degrees are offered in the College of Engineering at Notre Dame ? \\
			\midrule[0.5pt]
			\textbf{\textsc{Elmer}-Hard} & What is the name of Frederick Binaryinary at St. Joseph ? & How many B.S degrees offered at the College of the engineering ? \\
			\midrule[0.5pt]
			\textbf{\textsc{Elmer}-Soft}  & What is the name of the main seminary at St. Joseph ? & How many B.S. degrees offered at the College of departments ? \\
			\bottomrule[1pt]
		\end{tabular}
	\caption{Qualitative examples on SQuAD v1.1 dataset.} 
	\label{tab:squad}
\end{table*}

\begin{table*}[t]
\small
\centering
	\begin{tabular}{l | p{0.41\textwidth} | p{0.41\textwidth}}
			\toprule[1pt]
			\textbf{Histories} &  i love to meet new people . i have a turtle named timothy . my favorite sport is ultimate frisbee . my parents are living in bora bora . autumn is my favorite season . [SEP] hello , how are you doing tonight ? i am well an loving this interaction how are you ? i am great . i just got back from the club . & i just bought a brand new house . i like to dance at the club . i run a dog obedience school . i have a big sweet tooth . i like taking and posting selkies . [SEP] hello , how are you doing tonight ? i am well an loving this interaction how are you ? i am great . i just got back from the club . this is my favorite time of the year season wise i would rather eat chocolate cake during this season . what club did you go to ? me an timothy watched tv i went to club chino . what show are you watching ? lol oh okay kind of random do you live in a house or apartment ? we watched a show about animals like him i love those shows . i am really craving cake . why does that matter any ? i went outdoors to play frisbee \\
			\midrule[0.5pt]
			\textbf{Reponses} & this is my favorite time of the year season wise & it matters because i have a sweet tooth . \\
			\midrule[0.5pt]
			\textbf{\textsc{Elmer}-Hard} & what team of do you do ? ultimate & i like to exercise with my dogs . \\
			\midrule[0.5pt]
			\textbf{\textsc{Elmer}-Soft}  & i love club! do you have new friends ? & cool. i spend time with my dog . \\
			\bottomrule[1pt]
		\end{tabular}
	\caption{Qualitative examples on PersonaChat dataset.} 
	\label{tab:pc}
\end{table*}

\end{document}